\documentclass[runningheads]{llncs}

\usepackage{graphicx}
\usepackage{tikz}
\usepackage{comment}
\usepackage{amsmath,amssymb} 
\usepackage{color}
\usepackage{epsfig}
\usepackage{lineno}

\usepackage[breaklinks=true,bookmarks=false]{hyperref}

\newsavebox\CBox
\def\textBF#1{\sbox\CBox{#1}\resizebox{\wd\CBox}{\ht\CBox}{\textbf{#1}}}

\begin{document}
\pagestyle{headings}
\mainmatter
\def\ECCVSubNumber{101}  

\title{AIM 2020 Challenge on Learned Image Signal Processing Pipeline} 

\titlerunning{AIM 2020 Learned ISP Challenge}
%
\author{Andrey Ignatov \and Radu Timofte \and
Zhilu Zhang \and Ming Liu \and Haolin Wang \and Wangmeng Zuo \and Jiawei Zhang \and Ruimao Zhang \and Zhanglin Peng \and Sijie Ren \and 
Linhui Dai \and Xiaohong Liu \and Chengqi Li \and Jun Chen \and
Yuichi Ito \and
Bhavya Vasudeva \and Puneesh Deora \and Umapada Pal \and
Zhenyu Guo \and Yu Zhu \and Tian Liang \and Chenghua Li \and Cong Leng \and
Zhihong Pan \and Baopu Li \and
Byung-Hoon Kim \and Joonyoung Song \and Jong Chul Ye \and JaeHyun Baek \and
Magauiya Zhussip \and Yeskendir Koishekenov \and Hwechul Cho Ye \and
Xin Liu \and Xueying Hu \and Jun Jiang \and Jinwei Gu \and
Kai Li \and Pengliang Tan \and
Bingxin Hou $^*$
}

\institute{}
\authorrunning{A. Ignatov,  R. Timofte et al.}

\maketitle

\begin{abstract}
This paper reviews the second AIM learned ISP challenge and provides the description of the proposed solutions and results. The participating teams were solving a real-world RAW-to-RGB mapping problem, where to goal was to map the original low-quality RAW images captured by the Huawei P20 device to the same photos obtained with the Canon 5D DSLR camera. The considered task embraced a number of complex computer vision subtasks, such as image demosaicing, denoising, white balancing, color and contrast correction, demoireing, etc. The target metric used in this challenge combined fidelity scores (PSNR and SSIM) with solutions' perceptual results measured in a user study. The proposed solutions significantly improved the baseline results, defining the state-of-the-art for practical image signal processing pipeline modeling.
\end{abstract}

\section{Introduction}
\label{sec:introduction}

\let\thefootnote\relax\footnotetext{
$^*$ A. Ignatov and R. Timofte (\{andrey,radu.timofte\}@vision.ee.ethz.ch, ETH Zurich) are the challenge organizers, while the other authors participated in the challenge.\\
The Appendix~\ref{sec:affiliations} contains the authors' teams and affiliations.\\
AIM 2020 webpage: \url{https://data.vision.ee.ethz.ch/cvl/aim20/}}

Recently, the advent of deep learning, end-to-end learning paradigms, adversarial learning and the continuous improvements in memory and computational hardware led to tremendous advances in a number of research fields including computer vision, graphics, and computational photography. Particularly, the image restoration, enhancement and manipulation topics have witnessed an increased interest from the researchers, which resulted in an explosion of works defining and proposing novel solutions to improve different image quality aspects~\cite{Timofte_2018_CVPR_Workshops,Blau_2018_ECCV_Workshops,Abdelhamed_2019_CVPR_Workshops,Cai_2019_RealSR,gu2019brief,NTIRE_Dehazing_2019,Nah_2019_CVPR_Workshops,Gu_2019_CVPR_Workshops}, including its resolution, blur, noise, color rendition, perceptual quality, etc.
One of the most important real-world problems is the restoration and enhancement of the low-quality images recorded by compact camera sensors available in portable mobile devices~\cite{ignatov2017dslr,ignatov2018ai,ignatov2019ai,ignatov2018pirm,Chen_2018_CVPR} that are the prime source of media recordings nowadays.
In 2017, the first works were proposed to deal with a comprehensive image enhancement~\cite{ignatov2017dslr,ignatov2017wespe}. They were followed by a large number of subsequent papers that have significantly improved the baseline results~\cite{zhu2018range,stoutz2018fast,vu2018fast,hui2018perception,liu2018deep}. The PIRM challenge on perceptual image enhancement on smartphones~\cite{ignatov2018pirm}, the NTIRE 2019 challenge on image enhancement~\cite{ignatov2019ntire} that were working with a diverse DPED dataset~\cite{ignatov2017dslr} and several other NTIRE and AIM challenges were instrumental for producing a large number of efficient solutions and for further development in this field.

The AIM 2020 challenge on learned image signal processing pipeline is a step forward in benchmarking example-based single image enhancement. Same as the first RAW to RGB mapping challenge~\cite{ignatov2019aim}, it is targeted at processing and enhancing RAW photos obtained with small mobile camera sensors. AIM 2020 challenge uses a large-scale Zurich RAW to RGB (ZRR) dataset~\cite{ignatov2020replacing} consisting of RAW photos captured with the Huawei P20 mobile camera and the Canon 5D DSLR, and is taking into account both quantitative and qualitative visual results of the proposed solutions. In the next sections we describe the challenge and the corresponding dataset, present and discuss the results and describe the proposed methods.

This challenge is one of the AIM 2020 associated challenges on:
scene relighting and illumination estimation~\cite{elhelou2020aim_relighting}, image extreme inpainting~\cite{ntavelis2020aim_inpainting}, learned image signal processing pipeline~\cite{ignatov2020aim_ISP}, rendering realistic bokeh~\cite{ignatov2020aim_bokeh}, real image super-resolution~\cite{wei2020aim_realSR}, efficient super-resolution~\cite{zhang2020aim_efficientSR}, video temporal super-resolution~\cite{son2020aim_VTSR} and video extreme super-resolution~\cite{fuoli2020aim_VXSR}.

\section{AIM 2020 Challenge on Learned Image Signal Processing Pipeline}

\begin{figure*}[t!]
\centering
\setlength{\tabcolsep}{1pt}
\resizebox{\linewidth}{!}
{
\begin{tabular}{cccc}
\scriptsize{Huawei P20 RAW - Visualized}\normalsize & \scriptsize{Huawei P20 ISP}\normalsize & \scriptsize{Canon 5D Mark IV}\normalsize\\
    \includegraphics[width=0.33\linewidth]{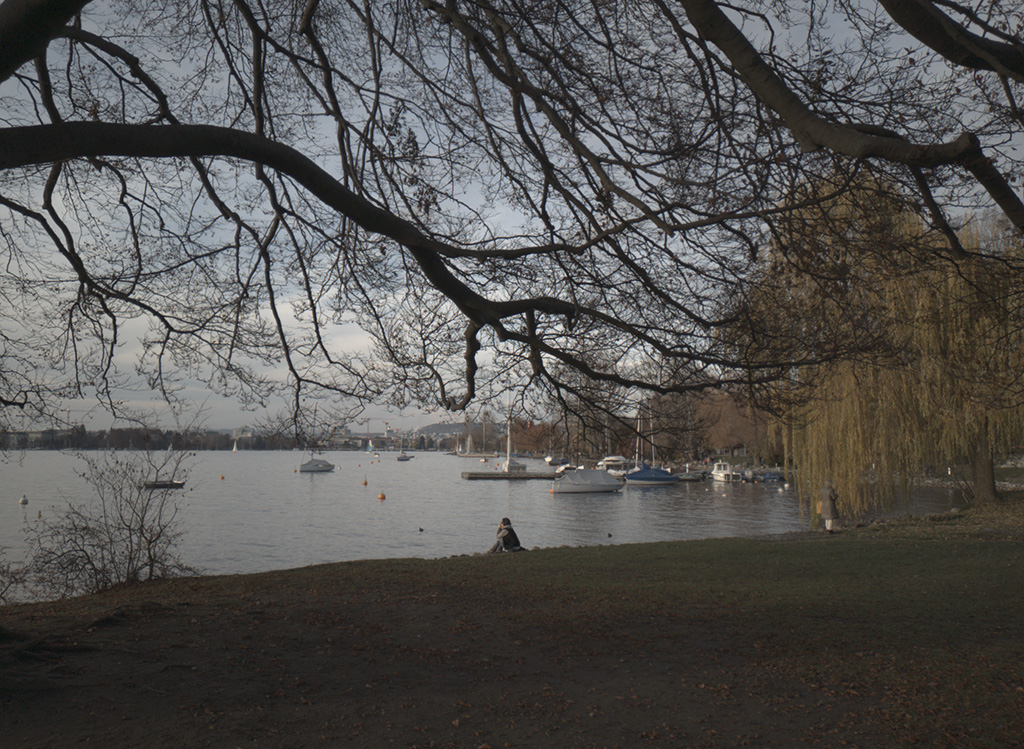}&
    \includegraphics[width=0.33\linewidth]{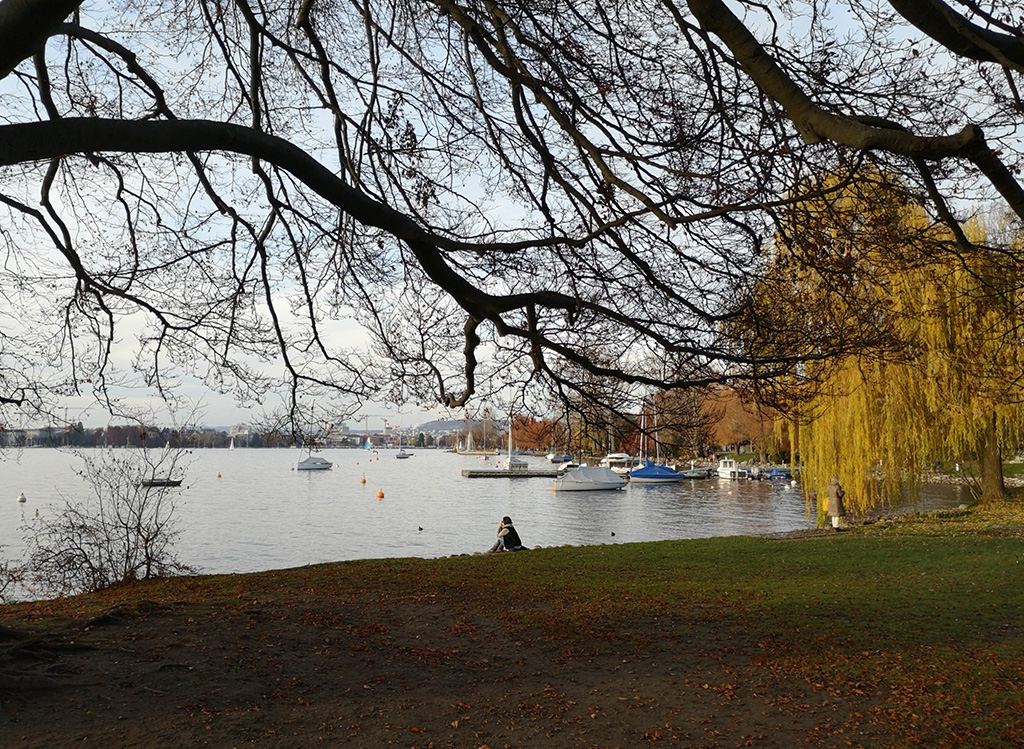}&
    \includegraphics[width=0.33\linewidth]{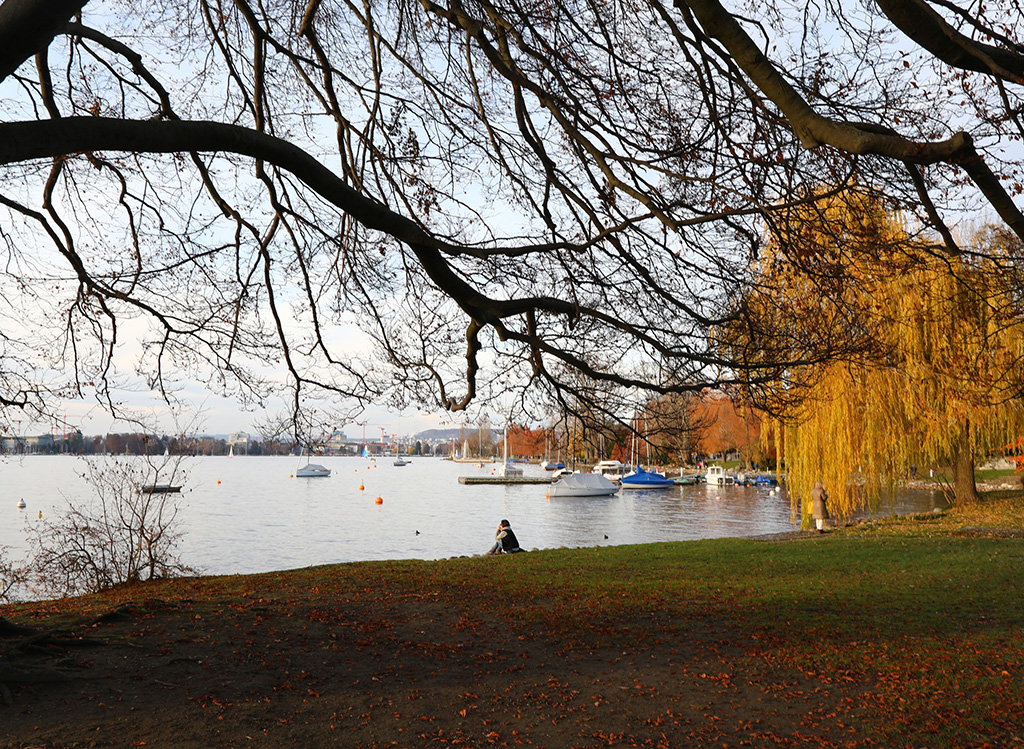}
\end{tabular}
}
\vspace{0.2cm}
\caption{Example set of images from the collected Zurich RAW to RGB dataset. From left to right: original RAW image visualized with a simple ISP script, RGB image obtained with P20's built-in ISP system, and Canon 5D Mark IV target photo.}
\label{fig:example_photos}
\end{figure*}

One of the biggest challenges in the RAW-to-RGB mapping task is to get high-quality real data that can be used for training deep models. To tackle this problem, we are using a large-scale ZRR dataset~\cite{ignatov2020replacing} dataset consisting of 20 thousand photos that was collected using Huawei P20 smartphone capturing RAW photos and a professional high-end Canon 5D Mark IV camera with Canon EF 24mm f/1.4L fast lens. RAW data was read from P20's 12.3 MP Sony Exmor IMX380 Bayer camera sensor~-- though this phone has a second 20 MP monochrome camera, it is only used by Huawei's internal ISP system, and the corresponding images cannot be retrieved with any public camera API. The photos were captured in automatic mode, and default settings were used throughout the whole collection procedure. The data was collected over several weeks in a variety of places and in various illumination and weather conditions. An example set of captured images is shown in Figure~\ref{fig:example_photos}.

Since the captured RAW--RGB image pairs are not perfectly aligned, we first performed their matching using the same procedure as in~\cite{ignatov2017dslr}. The images were first aligned globally using SIFT keypoints and RANSAC algorithm. Then, smaller patches of size 448$\times$448 were extracted from the preliminary matched images using a non-overlapping sliding window. Two windows were moving in parallel along the two images from each RAW-RGB pair, and the position of the window on DSLR image was additionally adjusted with small shifts and rotations to maximize the cross-correlation between the observed patches. Patches with cross-correlation less than 0.9 were not included into the dataset to avoid large displacements.
This procedure resulted in 48043 RAW-RGB image pairs (of size 448$\times$448$\times$1 and 448$\times$448$\times$3, respectively) that were later used for training / validation (46.8K) and testing (1.2K) the models. RAW image patches were additionally reshaped into the size of 224$\times$224$\times$4, where the four channels correspond to the four colors of the RGBG Bayer filer.
It should be mentioned that all alignment operations were performed only on RGB DSLR images, therefore RAW photos from Huawei P20 remained unmodified, containing the same values as were obtained from the camera sensor.

\subsection{Tracks and Competitions}

The challenge consists of the following phases:

\vspace{-0.8mm}
\begin{enumerate}
\setlength\itemsep{-0.2mm}
\item[i] \textit{development:} the participants get access to the data;
\item[ii] \textit{validation:} the participants have the opportunity to validate their solutions on the server and compare the results on the validation leaderboard;
\item[iii] \textit{test:} the participants submit their final results, models, and factsheets.
\end{enumerate}
\vspace{-0.8mm}

All submitted solutions were evaluated based on three measures:

\vspace{-0.8mm}
\begin{itemize}
\setlength\itemsep{-0.2mm}
\item PSNR measuring fidelity score,
\item SSIM, a proxy for perceptual score,
\item MOS scores measured in the user study for explicit image quality assessment.
\end{itemize}
\vspace{-0.8mm}

The AIM 2020 learned ISP pipeline challenge consists of two tracks. In the first ``Fidelity'' track, the target is to obtain an output image with the highest pixel fidelity to the ground truth as measured by PSNR and SSIM metrics. Since SSIM and PSNR scores are not reflecting many aspects of real image quality, in the second, ``Perceptual'' track, we are evaluating the solutions based on their Mean Opinion Scores (MOS). For this, we conduct a user study evaluating the visual results of all proposed methods. The users were asked to rate the quality of each submitted solution (based on 42 full resolution enhanced test images) by selecting one of five quality levels
(5 - comparable image quality, 4 - slightly worse, 3 - notably worse, 2 - poor image quality, 1 - completely corrupted image) for each method result in comparison with the original Canon images. The expressed preferences were averaged per each test image and then per each method to obtain the final MOS.

\section{Challenge Results}

\begin{table*}[tbh!]
\centering
\resizebox{\linewidth}{!}
{
\begin{tabular}{l|c|ccc|cc|ccc}
\multicolumn{2}{c|}{}&\multicolumn{3}{c|}{Factsheet Info}&\multicolumn{2}{c|}{Track 1: Fidelity}&\multicolumn{3}{c}{Track 2: Perceptual}\\

\hline
Team \, & \, Author \, & \, Framework \, & \, Hardware, GPU \, & \, Runtime, s \, & \, PSNR$\uparrow$ \, & \, SSIM$\uparrow$ & \, PSNR$\uparrow$ \, & \, SSIM$\uparrow$ \, & \, MOS$\uparrow$ \\
\hline
\hline
MW-ISPNet & \, zhangzhilu \, & PyTorch & 4 $\times$ GeForce GTX 1080 Ti & $\sim$1 & 21.91 & 0.7842 & 21.57 & 0.7770 & \textBF{4.7} \\
MacAI & \, itb202d \, & PyTorch & 2 $\times$ GeForce RTX 2080 Ti & 0.83 & 21.86 & 0.7807 & 21.86 & 0.7807 & 4.5 \\
Vermilion Vision & \, wataridori2010 \, & TensorFlow & GeForce RTX 2080 Ti & 0.062 & 21.40 & 0.7834 & 21.40 & 0.7834 & 4.2 \\
Eureka & \, bhavya\_vasudeva \, & Keras (TF) & 4 $\times$ GeForce GTX 1080 Ti & 0.078 & 21.18 & 0.7794 & 21.18 & 0.7794 & 4.1 \\
Airia\_CG & \, mo\_ming \, & PyTorch & 8 $\times$ Nvidia TITAN Xp & - & \textBF{22.26} & \textBF{0.7913} & 21.01 & 0.7691 & 4 \\
Baidu & \, zhihongp \, & PyTorch & GeForce GTX 2080 Ti & 1.2 & 21.91 & 0.7829 & \textBF{21.91} & 0.7829 & 4 \\
skyb & \, egyptdj \, & PyTorch & Nvidia Tesla V100 & 0.2 & 21.93 & 0.7865 & 21.73 & \textBF{0.7891} & 3.8 \\
STAIR & \, dark\_1im1ess \, & PyTorch & 4 $\times$ GeForce GTX 1080 & 0.59 & 21.57 & 0.7846 & 21.57 & 0.7846 & 3.5 \\
Sensebrainer & \, acehu \, & PyTorch & 4 $\times$ Nvidia Tesla V100 & 0.075 & 21.14 & 0.7729 & 21.14 & 0.7729 & 3.2 \\
bupt-mtc206 & \, TheClearwind \, & PyTorch & GeForce GTX 1080 Ti & 0.03 & 20.19 & 0.7622 & 20.19 & 0.7622 & 2.4 \\
BingSoda & \, houbingxin \, & PyTorch & Nvidia TITAN RTX & 0.04 & 20.14 & 0.7438 & 20.14 & 0.7438 & 2.2 \\
\end{tabular}
}
\vspace{2.6mm}
\caption{\small{AIM 2020 learned ISP pipeline challenge results and final rankings. The results are sorted based on the MOS scores.}}
\label{tab:results}
\end{table*}

The Track 1 of the challenge attracted more than 110 registered participants and the Track 2 more than 80. However, only 11 teams provided results in the final phase together with factsheets and codes for reproducibility.
Table~\ref{tab:results} summarizes the final test phase challenge results in terms of PSNR, SSIM and MOS scores for each submitted solution in the two tracks in addition to self-reported hardware / software configurations and runtimes. Short descriptions of the proposed solutions are provided in section~\ref{sec:solutions}, and the team details (contact email, members and affiliations) are listed in Appendix~\ref{sec:affiliations}.

\subsection{Architectures and Main Ideas}

All the proposed methods are relying on end-to-end deep learning-based solutions. The majority of submitted models have a multi-scale encoder-decoder architecture and are processing the images at several scales. This allows to introduce global image manipulations and to increase the training speed / decrease GPU RAM consumption as all heavy image processing is done on images of low resolution. Additionally, many challenge participants used channel-attention RCAN~\cite{zhang2018image} modules and various residual connections as well as discrete wavelet transform layers instead of the standard pooling ones to prevent the information loss. The majority of teams were using the MSE, $L1$, SSIM, VGG-based and color-based loss functions, while GAN loss was considered by only one team. Almost all participants are using Adam optimizer~\cite{kingma2014adam} to train deep learning models and PyTorch framework to implement and train the networks.

\subsection{Performance}

\paragraph{Quality.}
Team Airia\_CG achieves the best fidelity with the best PSNR and SSIM scores in Track 1, while team MW-ISPNet is the winner of the Track 2, achieving the best perceptual quality measured by Mean Opinion Scores (MOS) through a user study.

\bigskip

\noindent Airia\_CG, MW-ISPNet and skyb are the only teams that submitted different solutions to the two tracks. Second to MW-ISPNet's 22.26dB PSNR, there are four teams with similar results in the range of [21.86-21.93]. We also note that there is a good correlation between PSNR / SSIM scores and the perceptual ranking: the poorest perceptual quality is achieved by solutions that have the lowest PSNR and SSIM scores. Notable exceptions are the solutions proposed by Eureka and Airia\_CG for track 2 with good perceptual quality but poor PSNR. MacAI provides a solution with a good balance between between fidelity (21.86dB PSNR) and perceptual quality (4.5 MOS, second only to MW-ISPNet).

\paragraph{Runtime.}
The best fidelity and the best perceptual quality winning solutions are also among the more computational demanding in this challenge. MW-ISPNet reports around 1s per image crop. At the same time, order of magnitude faster solutions rank at the bottom in both tracks.

\subsection{Discussion}

The AIM 2020 challenge on learned image signal processing (ISP) pipeline promoted a novel direction of research aiming at replacing the current tedious and expensive handcrafted ISP solutions with data-driven learned ones capable to surpass them in terms of image quality. For this purpose, the participants were asked to map the smartphone camera RAW images not to the RGB outputs produced by a commercial handcrafted ISP but to the higher quality images captured with a high-end DSLR camera. The challenge employed the ZRR dataset~\cite{ignatov2020replacing} containing paired and aligned photos captured with the Huawei P20 smartphone and Canon 5D Mark IV DSLR camera. Many of the proposed approaches significantly improved over the original RAW images in perceptual quality in the direction of the DSLR quality target. The challenge through the proposed solutions defines the state-of-the-art for the practical learned ISP \textit{aka} RAW to RGB image mapping task.


\section{Challenge Methods and Teams}
\label{sec:solutions}

This section describes solutions submitted by all teams participating in the final stage of the AIM 2020 challenge on learned ISP pipeline.

\smallskip

\subsection{MW-ISPNet}

\begin{figure}[h!]
\centering
\resizebox{1.0\linewidth}{!}
{
\includegraphics[width=1.0\linewidth]{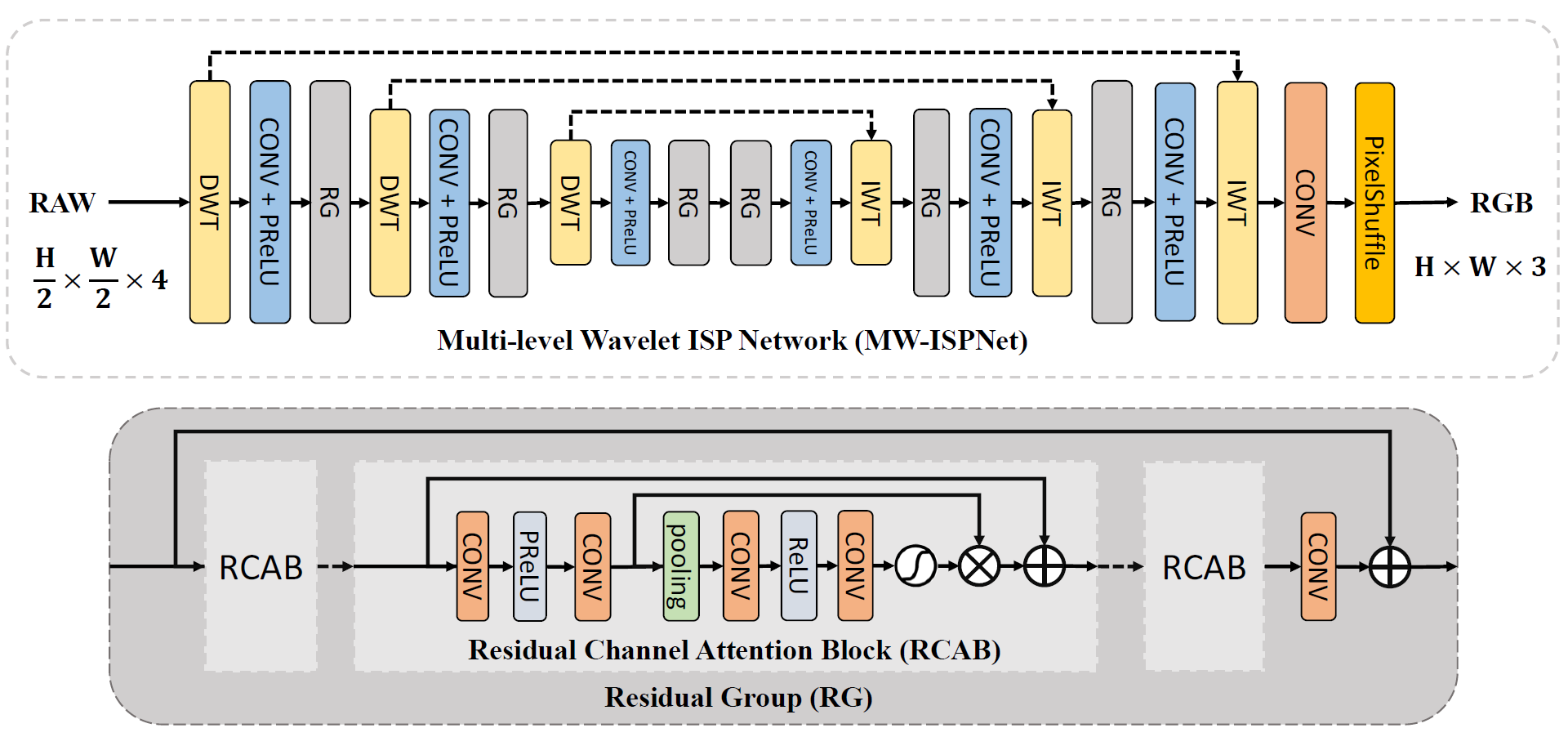}
}
\caption{\small{MW-ISPNet multi-level wavelet network.}}
\label{fig:MWISPNet}
\end{figure}

Team MW-ISPNet proposed a U-Net based multi-level wavelet ISP network (MW-ISPNet) illustrated in Fig.~\ref{fig:MWISPNet} that takes advantage of the MWCNN~\cite{liu2018multi} and RCAN~\cite{zhang2018image} architectures. In each U-Net level of this model, a residual group (RG) composed of 20 residual channel attention blocks (RCAB) is embedded. The standard downsampling and upsampling operations are replaced with a discrete wavelet transform based (DWT) decomposition to minimize the information loss in these layers.

The model is trained with a combination of the $L_1$, SSIM and VGG-based loss functions using the Adam algorithm. In the fidelity track, the authors used an additional MW-ISPNet model trained on the SIDD~\cite{abdelhamed2018high} dataset to perform raw image denoising, which outputs were passed to the main MW-ISPNet model. In the perceptual track, the authors added an adversarial loss following the LSGAN~\cite{mao2017least} paper to improve the perceptual quality of the produced images. Finally, the authors used a self-ensemble method averaging the eight outputs from the same model that is taking flipped and rotated images as an input.

\subsection{MacAI}

\begin{figure}[h!]
\centering
\resizebox{1.0\linewidth}{!}
{
\includegraphics[width=1.0\linewidth]{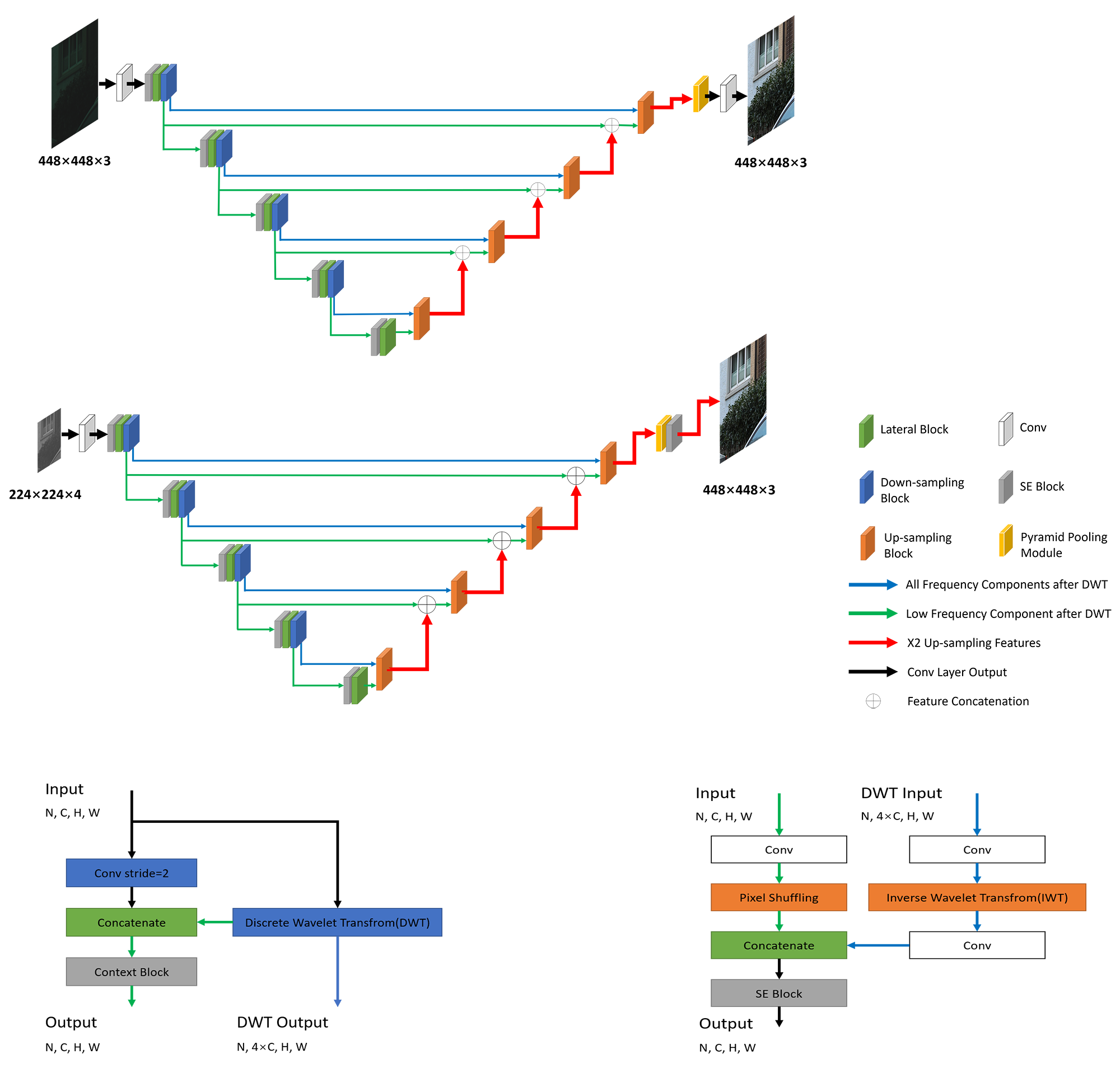}
}
\caption{\small{AWNet model proposed by team MacAI.}}
\label{fig:MacAI}
\end{figure}

Team MacAI presented the AWNet model~\cite{dai2020awnet} (Fig.~\ref{fig:MacAI}) utilizing the attention mechanism and wavelet transform and consisting of three blocks: lateral block, upsampling and downsampling blocks. The lateral block consists of several residual dense blocks (RDB) and a global context block (GCB)~\cite{cao2019gcnet}. Same as the previous team, the authors used the discrete wavelet transform (DWT) instead of the pooling layers to preserve the low-frequency information, though they additionally used the standard downscaling convolutional and pixel shuffle layers in parallel with the DWT layers to get a richer set of learned features. Finally, the authors trained one additional model that is taking a simple demosaiced raw image as an input (instead of the four Bayer channels), and combined the outputs of both models to produce the final image.

The model was trained with a combination of the Charbonnier, SSIM and VGG-based loss functions. The parameters of the model were optimized using the Adam algorithm with the initial learning rate of $1e-4$ halved every 10 epochs. A self-ensemble method averaging the eight outputs from the same model was used.

\subsection{Vermilion Vision}

\begin{figure}[h!]
\centering
\resizebox{1.0\linewidth}{!}
{
\includegraphics[width=1.0\linewidth]{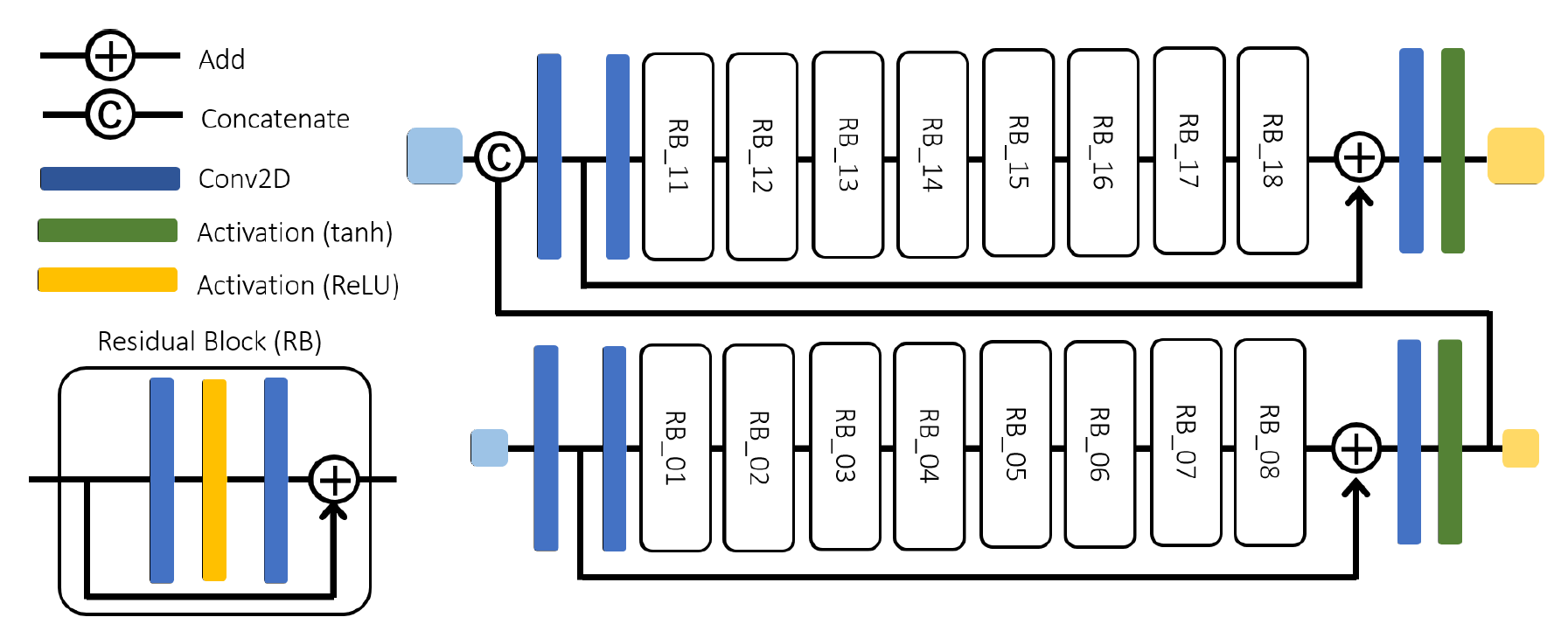}
}
\caption{\small{Vermilion Vision model architecture based on the Scale-recurrent Networks~\cite{tao2018scale}.}}
\label{fig:Vermilion}
\end{figure}

Vermilion Vision based their solution on Scale-Recurrent Networks~\cite{tao2018scale}. The architecture of the proposed model is shown in Fig.~\ref{fig:Vermilion}: it consists of 4 scales (though only two scales are shown here), \textit{tanh} activation function is used after the last layer to achieve a better tone-mapping effect. The authors used demosaiced (with a conventional DDFAPD algorithm) images as an input to their model, and the network was trained to minimize the standard MSE loss function. To get the final results, the authors were additionally averaging the outputs of the best 3 models.

\subsection{Eureka}

\begin{figure}[h!]
\centering
\resizebox{1.0\linewidth}{!}
{
\includegraphics[width=1.0\linewidth]{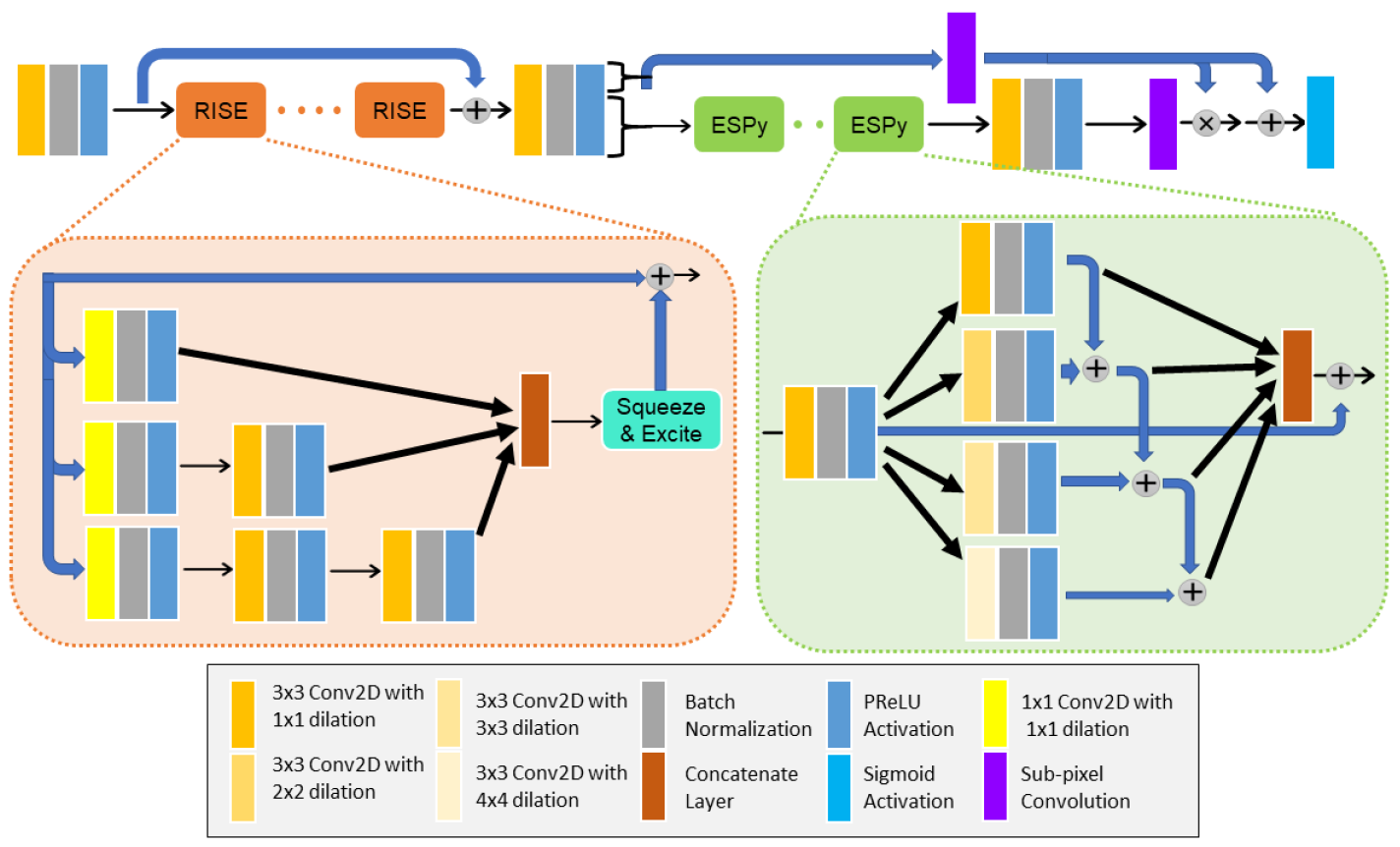}
}
\caption{\small{Network architecture with RISE and ESPy modules proposed by team Eureka.}}
\label{fig:Eureka}
\end{figure}

The solution proposed by Eureka (Fig.~\ref{fig:Eureka}) is using a residual inception module with squeeze and excite (RISE) module and an efficient spatial pyramid (ESPy) module. The first one has three parallel paths for convolutional layers with different receptive fields and is primarily focused on low-level operations. The features produced in this module are concatenated and passed through a squeeze block and excite block that controls the importance / weight of each channel. The second module is targeted at high-level enhancement and has four parallel paths with dilated convolutional layers and different dilation rates to cover larger image areas.

The output obtained from the sequence of RISE modules is divided into two parts: the first 12 channels are passed through the sub-pixel convolution layer to obtain the RGB channels, and the other channels are passed through the sequence of the ESPy modules. The final output is obtained as a linear regression of the aforementioned RGB channels, which weights are given by the output of the final ESPy module. The network is trained to minimize a combination of the following five losses: the mean absolute error (MAE), color loss (measured as the cosine distance between the RGB vectors), SSIM, VGG-based and exposure fusion~\cite{mertens2009exposure} loss. The parameters of the model are optimized using the Adam algorithm for 400 epochs.

\subsection{Airia\_CG}

\begin{figure}[h!]
\centering
\resizebox{1.0\linewidth}{!}
{
\includegraphics[width=1.0\linewidth]{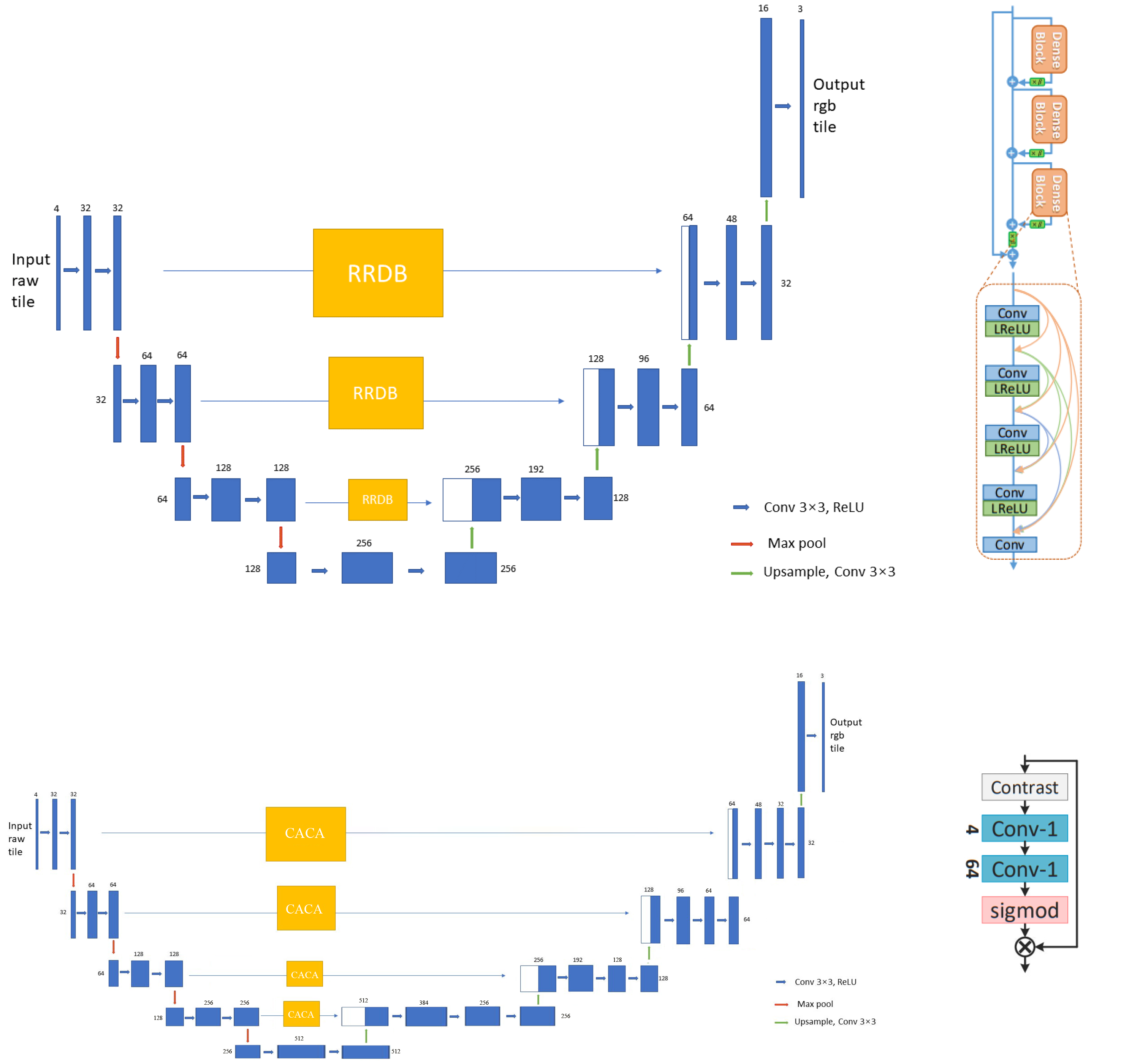}
}
\caption{\small{Progressive U-Net and EEDNetv2 architectures proposed by Airia\_CG.}}
\label{fig:Airia}
\end{figure}

Team Airia\_CG used two different approaches in this challenge. In the Perceptual track, it proposed a Progressive U-Net (PU-Net) architecture (Fig.~\ref{fig:Airia}, bottom) that is essentially a U-Net model augmented with Contrast-Aware Channel Attention modules~\cite{hui2019lightweight}, switchable normalization layers~\cite{luo2018differentiable} and pixel shuffle layers for upsampling the images. The authors have additionally cleaned the provided ZRR dataset by removing all blurred photos, and used the obtained image subset for training the model.

In the fidelity track, the authors used an ensemble of six different models: the PU-Net model described above, PyNET~\cite{ignatov2020replacing} and four EEDNets~\cite{zhu2020eednet} (Fig.~\ref{fig:Airia}, top) with slightly different architectures: EEDNetv1 uses simple copy and crop rather than RRDB modules compared to EEDNetv2, and EEDNetv4 adds a contrast-aware channel attention module after the RRDB module. The considered ensemble was able to improve the PSNR on the validation dataset by 0.61dB compared to its best single model.

\subsection{Baidu Research Vision}

\begin{figure}[h!]
\centering
\resizebox{1.0\linewidth}{!}
{
\includegraphics[width=1.0\linewidth]{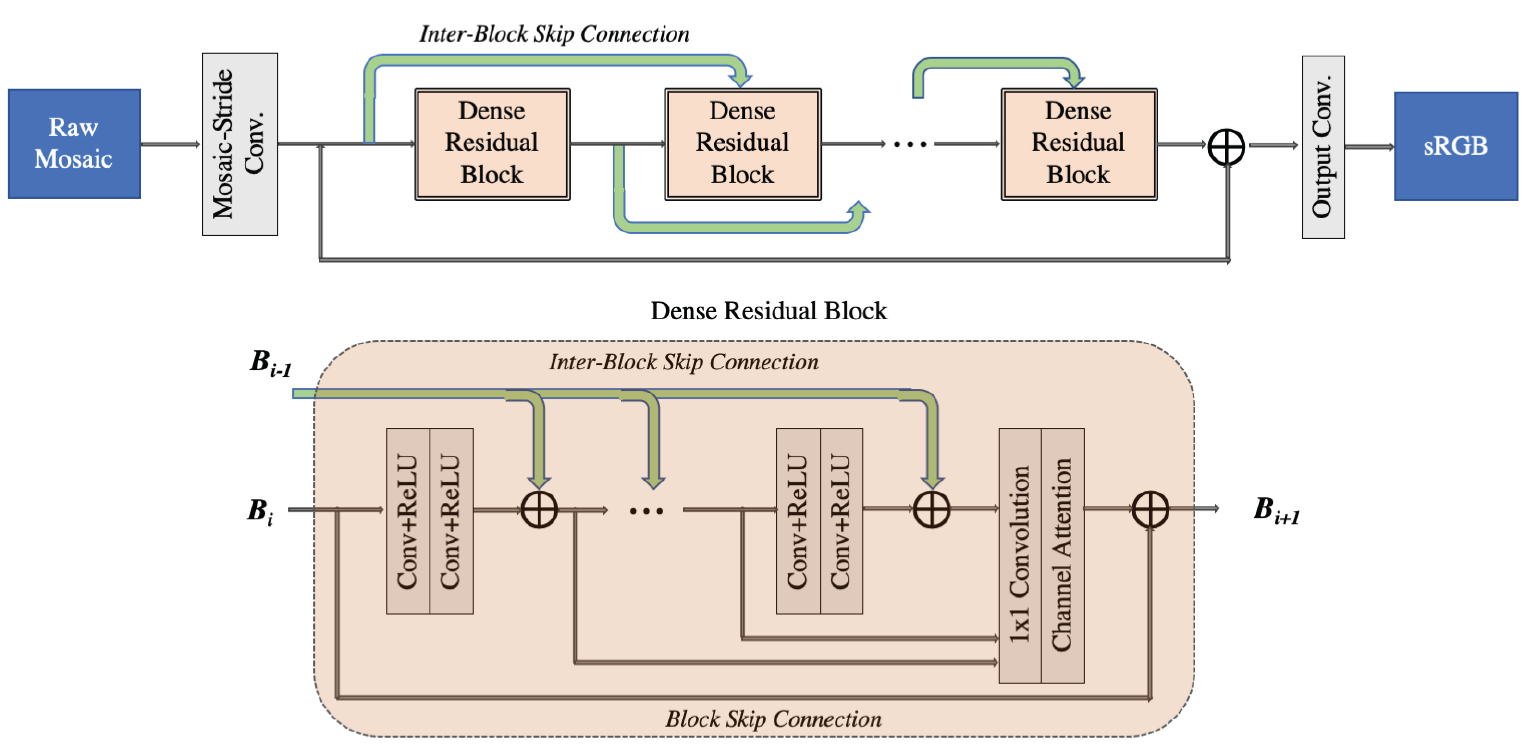}
}
\caption{\small{Mosaic-adaptive dense residual network proposed by Baidu.}}
\label{fig:Baidu}
\end{figure}

Baidu team based their solution on a mosaic-adaptive dense residual network (Fig.~\ref{fig:Baidu}). A mosaic stride convolution layer at the beginning of the model is used to extract mosaic-adaptive shallow features. The model is enhanced with additional channel-attention modules as in RCAN~\cite{zhang2018image}. The network was trained with a combination of the $L_1$ and SSIM loss functions, a self-assemble strategy was additionally used for generating the final results.

\subsection{Skyb}

\begin{figure}[h!]
\centering
\resizebox{1.0\linewidth}{!}
{
\includegraphics[width=1.0\linewidth]{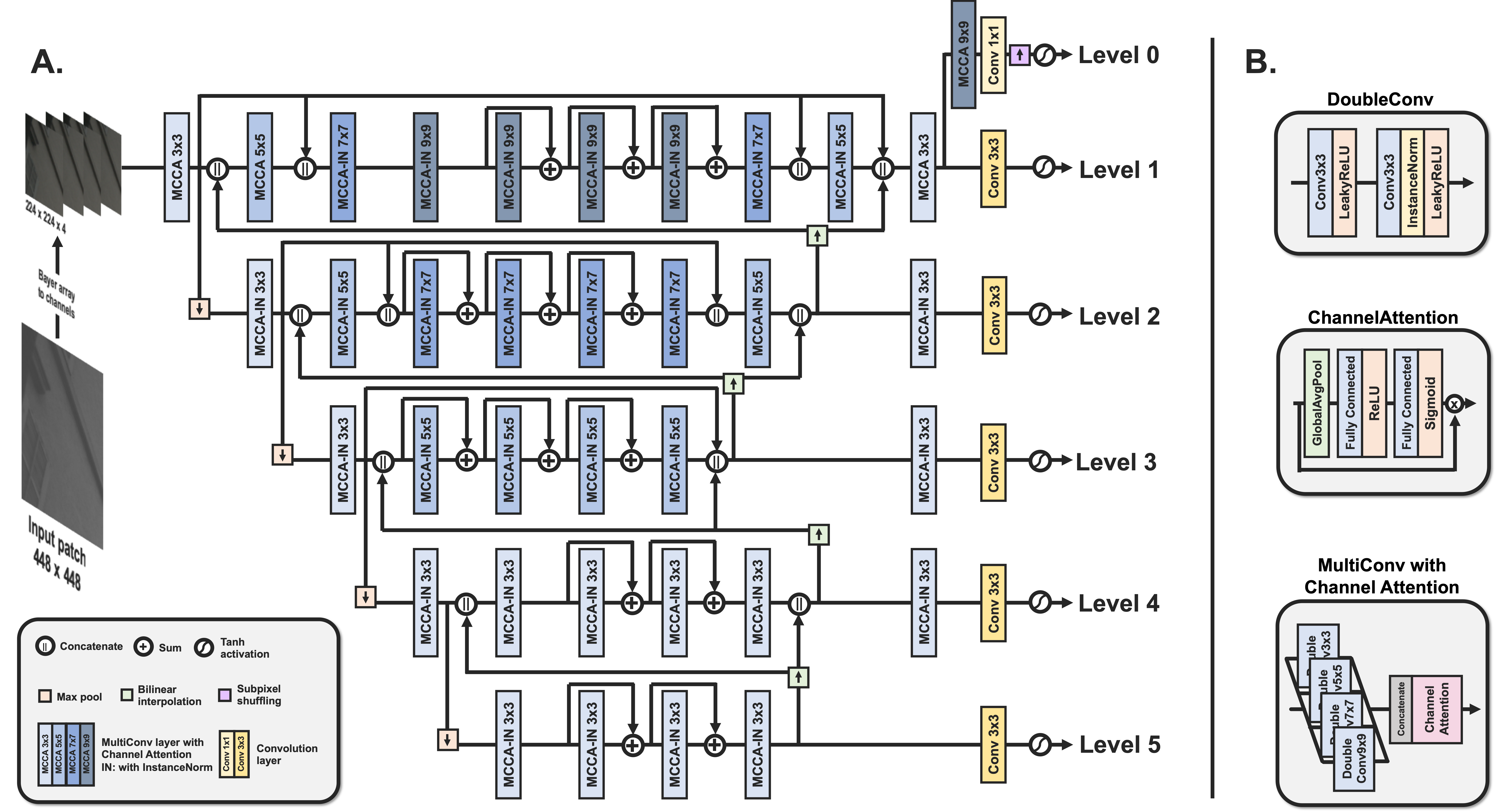}
}
\caption{\small{PyNet-CA model proposed by team Skyb.}}
\label{fig:Skyb}
\end{figure}

Skyb presented a PyNet-CA model~\cite{kim2020pynet-ca} (Fig.~\ref{fig:Skyb}) that adds several enhancements on top of the standard PyNET~\cite{ignatov2020replacing} architecture. In particular, the authors added RCAN style channel attention~\cite{zhang2018image} on top of the outputs from multi-convolutional layers. Besides that, some instance normalization ops were removed, an additional multi-convolutional layer was used for upscaling the final image, and a different one-cycle learning rate policy~\cite{smith2019super} was used for training each level of the model. The network was trained with a combination of the MSE, VGG-based and SSIM loss function taken in different combinations depending on the track and PyNET level, a self-ensemble strategy was additionally applied for producing the final outputs.

\subsection{STAIR}

\begin{figure}[h!]
\centering
\resizebox{1.0\linewidth}{!}
{
\includegraphics[width=0.9\linewidth]{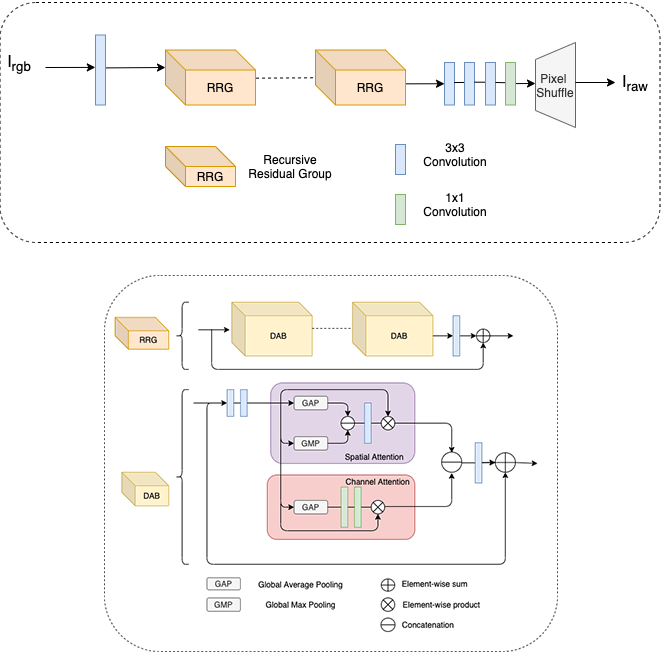}
}
\caption{\small{RRGNet model used by team STAIR.}}
\label{fig:STAIR}
\end{figure}

STAIR used the RRGNet model presented in Fig.~\ref{fig:STAIR} for restoring RGB images. The network utilizes a combination of spatial attention (SA) and channel attention (CA) blocks enhanced with residual connections between CNN modules. The model was trained to minimize $L_1$ loss, its parameters were optimized using the Adam algorithm for 30 epochs, a self-ensemble strategy was additionally applied for producing the final outputs.

\subsection{SenseBrainer}

\begin{figure}[h!]
\centering
\resizebox{1.0\linewidth}{!}
{
\includegraphics[width=1.0\linewidth]{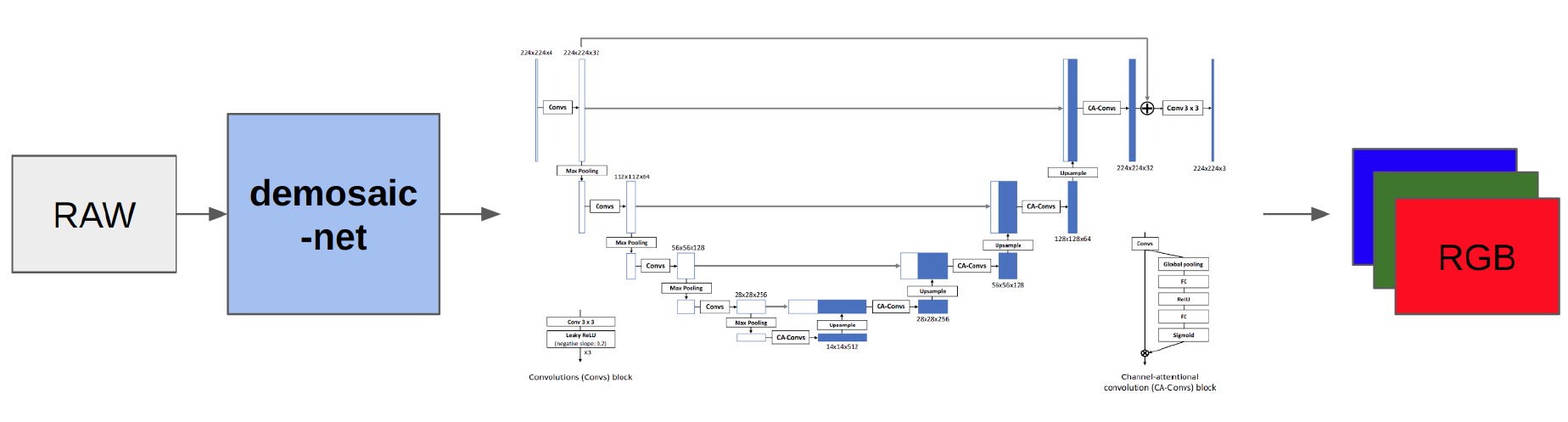}
}
\caption{\small{A Multiscaled U-Net architecture proposed by SenseBrainer.}}
\label{fig:SenseBrainer}
\end{figure}

SenseBrainer proposed a Multiscaled U-Net model displayed in Fig.~\ref{fig:SenseBrainer} for the considered task. The authors first processed raw images with the Demosaic-Net to produce RGB images reconstructed without color correction. Multiscaled U-Net was then trained to restore image colors with a combination of the MSE, color similarity and VGG-based losses, which weights were depending on the training scale. The authors have additionally removed the corrupted / misaligned images from the dataset to make the training more robust.

\subsection{Bupt-mtc206}

\begin{figure}[h!]
\centering
\resizebox{1.0\linewidth}{!}
{
\includegraphics[width=1.0\linewidth]{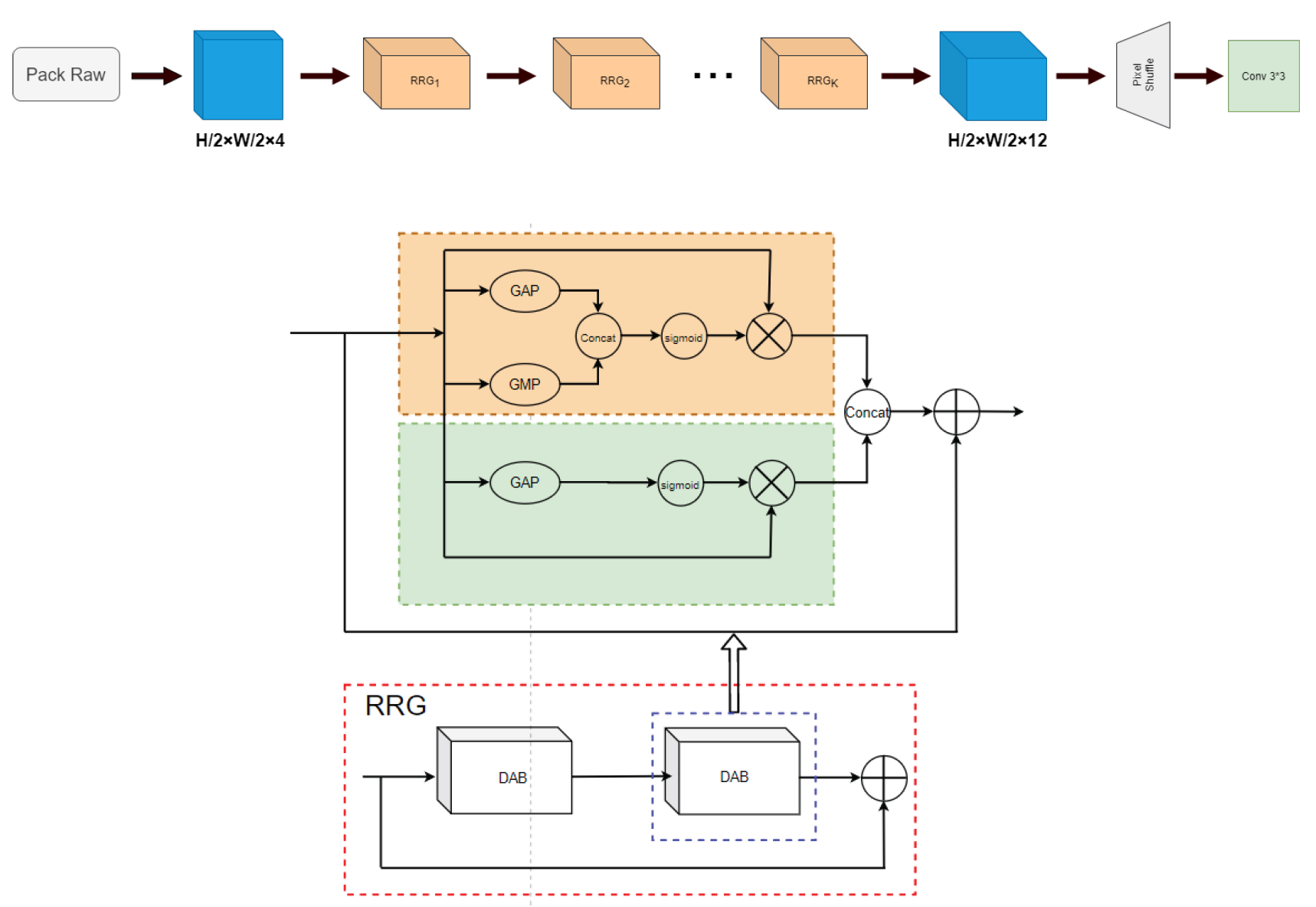}
}
\caption{\small{RRGNet model used by team Bupt-mtc206.}}
\label{fig:Bupt}
\end{figure}

Similarly to team STAIR, the authors tried to apply the RRGNet model (Fig.~\ref{fig:Bupt}) to the considered problem. Due to a hardware problem, the authors were not able to run their solution on full-resolution images, therefore they had to use the stitching method in their final submission.

\subsection{BingSoda}

\begin{figure}[h!]
\centering
\resizebox{1.0\linewidth}{!}
{
\includegraphics[width=1.0\linewidth]{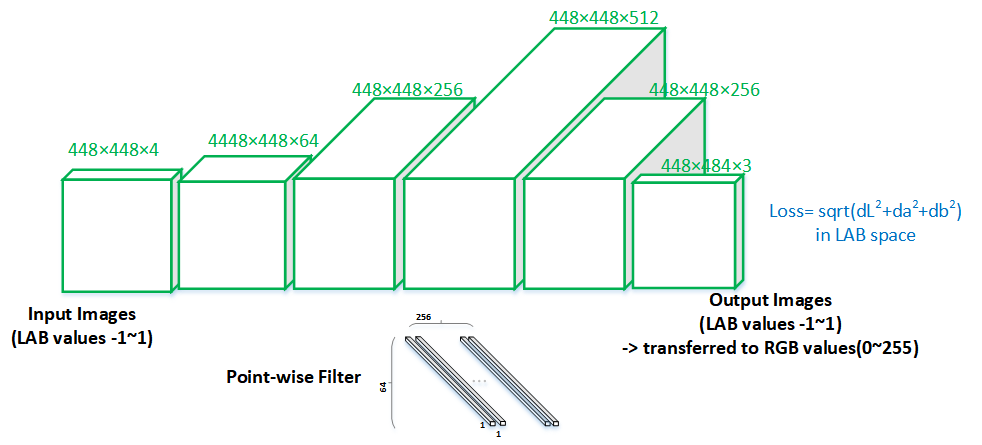}
}
\caption{\small{PWCD model proposed by team BingSoda.}}
\label{fig:BingSoda}
\end{figure}

Team BingSoda presented a Pixel-Wise Color Distance (PWCD) model illustrated in Fig.~\ref{fig:BingSoda}. The model was using the LAB color space instead of the RGB one, and was trained to minimize the CIELAB color difference between the predicted and target images.

\section*{Acknowledgments}
We thank the AIM 2020 sponsors: Huawei, MediaTek, Qualcomm, NVIDIA, Google and Computer Vision Lab / ETH Z\"urich.


\appendix
\section{Appendix 1: Teams and affiliations}
\label{sec:affiliations}

\bigskip


\textbf{AIM 2020 Learned ISP Challenge Team}
\small
\bigskip

\noindent
\textBF{Title:} AIM 2020 Challenge on Learned Image Signal Processing Pipeline

\smallskip

\noindent
\textBF{Members:} \hspace{0.7mm} Andrey Ignatov \,--\, \footnotesize{andrey@vision.ee.ethz.ch}, \small

\hspace{10.5mm} Radu Timofte \,--\, \footnotesize{radu.timofte@vision.ee.ethz.ch}

\smallskip

\noindent
\textBF{Affiliations:} \hspace{0.7mm} Computer Vision Lab, ETH Zurich, Switzerland


\bigskip

\smallskip

\noindent
\textbf{MW-ISPNet}
\small
\bigskip

\noindent
\textBF{Title:} Multi-level Wavelet ISP Network

\smallskip

\noindent
\textBF{Members:} \hspace{0.7mm} Zhilu Zhang $^1$ \,--\, \footnotesize{cszlzhang@outlook.com}, \small

\hspace{10.5mm} Ming Liu $^1$, Haolin Wang $^1$, Wangmeng Zuo $^1$

\hspace{10.5mm} Jiawei Zhang $^2$, Ruimao Zhang $^2$, Zhanglin Peng $^2$, Sijie Ren $^2$

\smallskip

\noindent
\textBF{Affiliations:} \hspace{0.7mm} $^1$~-- Harbin Institute of Technology, China

\hspace{12.9mm} $^2$~-- SenseTime, China

\bigskip

\smallskip
\noindent
\textbf{MacAI}
\small
\bigskip

\noindent
\textBF{Title:} Attentive Wavelet Network for Image ISP~\cite{dai2020awnet}

\smallskip

\noindent
\textBF{Members:} \hspace{0.7mm} Linhui Dai \,--\, \footnotesize{dail5@mcmaster.ca}, \small

\hspace{10.5mm} Xiaohong Liu, Chengqi Li, Jun Chen

\smallskip

\noindent
\textBF{Affiliations:} \hspace{0.7mm} McMaster University, Canada

\bigskip

\smallskip

\noindent
\textbf{Vermilion Vision}
\small
\bigskip

\noindent
\textBF{Title:} Scale Recurrent Deep Tone Mapping

\smallskip

\noindent
\textBF{Members:} \hspace{0.7mm} Yuichi Ito \,--\, \footnotesize{yito@vermilionvision.net}, \small

\smallskip

\noindent
\textBF{Affiliations:} \hspace{0.7mm} Vermilion Vision, United States

\bigskip

\smallskip

\noindent
\textbf{Eureka}
\small
\bigskip

\noindent
\textBF{Title:} Local and Global Enhancement Network as Learned ISP

\smallskip

\noindent
\textBF{Members:} \hspace{0.7mm} Bhavya Vasudeva \,--\, \footnotesize{bhavyavasudeva10@gmail.com}, \small

\hspace{10.5mm} Puneesh Deora, Umapada Pal

\smallskip

\noindent
\textBF{Affiliations:} \hspace{0.7mm} CVPR Unit, ISI Kolkata, India

\bigskip

\smallskip

\noindent
\textbf{Airia\_CG}
\small
\bigskip

\noindent
\textBF{Title 1:} EEDNet: Enhanced Encoder-Decoder Network

\smallskip

\noindent
\textBF{Title 2:} PUNet: Progressive U-Net via Contrast-aware
Channel Attention

\smallskip

\noindent
\textBF{Members:} \hspace{0.7mm} Zhenyu Guo \,--\, \footnotesize{guozhenyu2019@ia.ac.cn}, \small

\hspace{10.5mm} Yu Zhu, Tian Liang, Chenghua Li, Cong Leng

\smallskip

\noindent
\textBF{Affiliations:} \hspace{0.7mm} Nanjing Artificial Intelligence Chip Research, Institute of Automation

\hspace{12.9mm} Chinese Academy of Sciences (AiRiA), MAICRO, China

\bigskip

\smallskip

\noindent
\textbf{Baidu Research Vision}
\small
\bigskip

\noindent
\textBF{Title:} Learned Smartphone ISP using Mosaic-Adaptive Dense Residual Network

\smallskip

\noindent
\textBF{Members:} \hspace{0.7mm} Zhihong Pan \,--\, \footnotesize{zhihongpan@baidu.com}, \small

\hspace{10.5mm} Baopu Li

\smallskip

\noindent
\textBF{Affiliations:} \hspace{0.7mm} Baidu Research, United States

\bigskip

\smallskip

\noindent
\textbf{Skyb}
\small
\bigskip

\noindent
\textBF{Title:} PyNet-CA: Enhanced PyNet with Channel Attention for Mobile ISP

\smallskip

\noindent
\textBF{Members:} \hspace{0.7mm} Byung-Hoon Kim $^1$ \,--\, \footnotesize{egyptdj@kaist.ac.kr}, \small

\hspace{10.5mm} Joonyoung Song $^1$, Jong Chul Ye $^1$, JaeHyun Baek $^2$

\smallskip

\noindent
\textBF{Affiliations:} \hspace{0.7mm} $^1$~--  Korea Advanced Institute of Science and Technology (KAIST),

\hspace{12.9mm} $^2$~-- Amazon Web Services, South Korea

\bigskip

\smallskip

\noindent
\textbf{STAIR}
\small
\bigskip

\noindent
\textBF{Title:} Recursive Residual Group Network for Image Mapping

\smallskip

\noindent
\textBF{Members:} \hspace{0.7mm} Magauiya Zhussip $^1$ \,--\, \footnotesize{magauiya173@gmail.com}, \small

\hspace{10.5mm} Yeskendir Koishekenov $^2$, Hwechul Cho Ye $^1$

\smallskip

\noindent
\textBF{Affiliations:} \hspace{0.7mm} $^1$~--  ST Unitas AI Research (STAIR), South Korea

\hspace{12.9mm} $^2$~-- Allganize, South Korea

\bigskip

\smallskip

\noindent
\textbf{SenseBrainer}
\small
\bigskip

\noindent
\textBF{Title:} Multiscaled UNet

\smallskip

\noindent
\textBF{Members:} \hspace{0.7mm} Xin Liu \,--\, \footnotesize{liuxin@sensebrain.site}, \small

\hspace{10.5mm} Xueying Hu, Jun Jiang, Jinwei Gu

\smallskip

\noindent
\textBF{Affiliations:} \hspace{0.7mm} SenseBrain, United States

\bigskip

\smallskip

\noindent
\textbf{Bupt-mtc206}
\small
\bigskip

\noindent
\textBF{Title:} RRGNet for Smartphone ISP

\smallskip

\noindent
\textBF{Members:} \hspace{0.7mm} Kai Li \,--\, \footnotesize{492071523@qq.com}, \small

\hspace{10.5mm} Pengliang Tan

\smallskip

\noindent
\textBF{Affiliations:} \hspace{0.7mm} Beijing University of Posts and Telecommunications, China

\bigskip

\smallskip

\noindent
\textbf{BingSoda}
\small
\bigskip

\noindent
\textBF{Title:} Pixel-Wise Color Distance (PWCD model)

\smallskip

\noindent
\textBF{Members:} \hspace{0.7mm} Bingxin Hou \,--\, \footnotesize{houbingxin@gmail.com}, \small

\smallskip

\noindent
\textBF{Affiliations:} \hspace{0.7mm} Santa Clara University, United States

\bigskip

\smallskip


{\small
\bibliographystyle{splncs04}
\bibliography{egbib}
}

\end{document}